\documentclass{article}

\usepackage{PRIMEarxiv}

\usepackage[utf8]{inputenc} 
\usepackage[T1]{fontenc}    
\usepackage{hyperref}       
\usepackage{url}            
\usepackage{booktabs}       
\usepackage{amsfonts}       
\usepackage{nicefrac}       
\usepackage{microtype}      
\usepackage{lipsum}
\usepackage{fancyhdr}       
\usepackage{graphicx}       
\graphicspath{{media/}}     
\usepackage{subcaption}
\usepackage{amsmath}
\title{Detecting Stylistic Fingerprints of Large Language Models}

\author{%
  Yehonatan Bitton, Elad Bitton, and Shai Nisan\thanks{Direct correspondence to: \{ shaini \} @ copyleaks.com.}\\[1ex]
  \small Copyleaks
}

\begin{document}
\maketitle

\begin{abstract}
Large language models (LLMs) have distinct and consistent stylistic fingerprints, even when prompted to write in different writing styles. Detecting these fingerprints is important for many reasons, among them protecting intellectual property, ensuring transparency regarding AI-generated content, and preventing the misuse of AI technologies. In this paper, we present a novel method to classify texts based on the stylistic fingerprints of the models that generated them. We introduce an LLM-detection ensemble that is composed of three classifiers with varied architectures and training data. This ensemble is trained to classify texts generated by four well-known LLM families: Claude, Gemini, Llama, and OpenAI. As this task is highly cost-sensitive and might have severe implications, we want to minimize false-positives and increase confidence. We consider a prediction as valid when all three classifiers in the ensemble unanimously agree on the output classification. Our ensemble is validated on a test set of texts generated by Claude, Gemini, Llama, and OpenAI models, and achieves extremely high precision (0.9988) and a very low false-positive rate (0.0004). Furthermore, we demonstrate the ensemble’s ability to distinguish between texts generated by seen and unseen models. This reveals interesting stylistic relationships between models. This approach to stylistic analysis has implications for verifying the originality of AI-generated texts and tracking the origins of model training techniques.
\end{abstract}


\section{Introduction}
\label{sec:headings}
\subsection{Motivation and Challenges}
Given the vast amounts of AI-generated content,\cite{brooks2024riseaigeneratedcontentwikipedia, AI-GeneratedContentandtheQuestionofCopyright} it is important to correctly attribute these texts to mitigate harmful content that may be generated by LLMs, protect intellectual property, and promote integrity, trust, and fairness in writing.\cite{abdali2024decodingaipentechniques, kwon2024comprehensive} 

However, the binary classification task (human vs. LLM) is insufficient; it is also important to identify exactly which LLM generated a given text.\cite{huang2025authorshipattributionerallms, Yang_Zhang_Chen_Zhang_Ma_Wang_Yu_2022} In this regard, we can treat the AI-generated text detection task as an authorship attribution task, extending the binary classification task to a multiclass classification task, where each class corresponds to a specific LLM.\cite{huang2025authorshipattributionerallms} 

Classifying which LLM wrote a text can enhance confidence in the detection outcomes, which is necessary since many AI detectors are 'black box' classifiers, meaning that their outputs are not easily explainable.\cite{Shi_2024} Correct LLM identification can promote greater integrity of the content, enhance authenticity of the information, as well as protect against more types of misuse of AI-generated content.\cite{Yang_Zhang_Chen_Zhang_Ma_Wang_Yu_2022} 

In particular, this ability can facilitate forensic and analytical investigations that are crucial to the AI market as a whole. For example, it can help in detecting hidden similarities between models, which may indicate instances of model distillation or unauthorized model reuse, such as when the output of one model is used to train another model. Some LLM vendors prohibit this kind of usage in certain scenarios,\footnote{For example, refer to the \href{https://openai.com/policies/row-terms-of-use/}{OpenAI Terms of Use} for more details on these restrictions (accessed February 6, 2025).} as it infringes on their intellectual property. 

\subsection{Related Work and Theoretical Background}
This research is built upon two main areas of study: authorship attribution and AI forensics. We explore the key theoretical aspects of each in turn.

\subsubsection{Authorship Attribution}
Authorship attribution is a widely researched field. It is a classification task, aimed at identifying an author of a document out of a given set of candidate authors.\cite{boenninghoff2019explainableauthorshipverificationsocial} It has many forensic applications, among them plagiarism detection, copyright infringement, authenticity verification, software analysis, and security attacks detection.\cite{info15030131} One main aspect in any authorship attribution attempt is the ability to quantify and capture a specific writing style of an author.\cite{stamatatos-2017-authorship}

Research in authorship attribution started in the late 19th century.\cite{sari2018neural} In the middle of the 20th century, early author attribution research attempted to capture and quantify writing style characteristics such as punctuation patterns, vocabulary richness, and word frequencies. To highlight unique stylistic patterns essential for determining authorship, researchers analyzed linguistic features and applied statistical measures using the new computational resources that became available during that time.\cite{puig2016unified} 

In the late 1990s, more comprehensive feature extraction techniques gained popularity. These enabled researchers to capture more stylistic elements such as n-grams and Part-of-Speech tags. New methods to process and analyze these features were used, especially machine learning methods such as KNN and Naive Bayes. These were found helpful in capturing even more subtle patterns in the writing style of authors, and have shown better accuracy and a greater ability to work with larger text corpora.\cite{uchendu-etal-2020-authorship, kumarage2023neuralauthorshipattributionstylometric, Authorship_attribution_Performance_of} During the 2010s, the usage of neural networks such as RNNs and CNNs further improved the accuracy in representing an author’s unique writing style.\cite{rhodes2015author, shrestha-etal-2017-convolutional} 

The advancements in transformer-based models in the late 2010s created a new type of author - AI writers - making it necessary to expand the author attribution task to include them. At that time, new approaches for authorship attribution began using classifiers based on pre-trained language models, an approach that proved especially accurate in detecting neural authors.\cite{huang2025authorshipattributionerallms, app13127255, kumarage2023neuralauthorshipattributionstylometric, guo2024benchmarkinglinguisticdiversitylarge}

\subsubsection{AI Forensics}
The task of detecting AI-generated text is a classification task. In most cases, it is usually treated as a binary classification problem - AI or human - but could also be a multi-class problem if one wishes to differentiate between specific AI text generators. The task becomes even more complex when a text is written with a combination of several AI models, or using AI and human authors, requiring the identification of each contributor.\cite{fraser2024detectingaigeneratedtextfactors}

In most cases, detecting AI-generated text is done using supervised learning frameworks. This requires labeled examples to train a classifier. The main metrics that are used are true-positive rates, true-negative rates, accuracy and F1 score. Classical approaches applied feature extraction on the texts in order to quantify stylistic or syntactic features such as perplexity, burstiness, and word density. These features are then fed into machine learning algorithms such as logistic regression or SVM (which provide an advantage of explainability), or into “deep” neural networks such as LSTMs. More modern approaches use pre-trained language models such as BERT, and fine-tune them for the specific classification task. This eliminates the need to manually define a list of features, and overall presents a much higher accuracy compared to feature-based methods, mainly because this approach is able to capture more general elements and nuances of the writing style, and is less dependent on a specific domain or topic.\cite{fraser2024detectingaigeneratedtextfactors, yadagiri-etal-2024-detecting, dathathri2024scalable, Kwon-Soonchan-and-Jang-Beakcheo}

Approaches that rely solely on capturing the differences between AI and human texts operate under the assumption that such differences exist; however, in reality, the differences between machine and human text may diminish over time as these systems evolve. A “watermarking” approach has recently emerged in an attempt to overcome these shortcomings. “Watermarks” can be added to the text during the generation phase (generative watermarking) or after it (edit-based watermarking).\cite{dathathri2024scalable} However, this approach has its own limitations, such as lower effectiveness on factual responses (as there are fewer options to control the output), and higher sensitivity to small edits of the watermarked text.\footnote{On the limitations of one of these methods, SynthID, please see \href{https://ai.google.dev/responsible/docs/safeguards/synthid}{Google’s SynthID documentation} (accessed February 6, 2025).}

\subsection{Bridging the Gap: Finding an LLM’s Fingerprints}
Research in linguistic analysis has already shown that LLMs have unique and consistent linguistic markers - “fingerprints” - which set them apart from each other, as well as from human-written texts. These fingerprints are consistent across domains, and persist even when the models are prompted to write in different writing styles. Their highly stable stylistic profile is due to the deterministic nature of their training, fine-tuning, and text generation processes.\cite{guo2024benchmarkinglinguisticdiversitylarge}

Models from the same family may share similar fingerprints (for example GPT4 and GPT3.5), but this isn’t always the case, as differences in training methods, architecture, fine-tuning, and generation techniques may lead to different writing styles even among models from the same family.\cite{mcgovern2024largelanguagemodelsleaving}

\subsection{Research Objectives and Motivation}
In this paper, we apply principles of authorship attribution combined with AI forensics strategies and techniques to develop a method for detecting and classifying specific LLM fingerprints. 

Our work aims to establish a highly reliable LLM-detection method that classifies texts based on the LLM family that generated them - Claude, Gemini, Llama, and OpenAI. This task has a high-stakes nature with significant consequences for wrongly identifying a text from one LLM as originating from another.\cite{The-Problem-with-False-Positives}

We therefore build upon the principles of the cost-sensitive classification paradigm,\cite{verbeke2021foundationscostsensitivecausalclassification} where the costs of different types of errors are not equal. In our case, the cost of incorrectly classifying an LLM as the writer of a certain text, involves a larger cost than failing to attribute a text to a certain LLM. Such costs include potential legal, reputational, and intellectual property implications of incorrect attribution.

\section{Approach}
\subsection{Ensemble Design and Rationale}
We build an ensemble of classifiers based on the principle that the combined decision of multiple different models is more accurate than that of a single model, given that the errors of the models are at least somewhat uncorrelated, and each individual model’s error rate is lower than 0.5.\cite{A-Survey-of-Ensemble-Learning}

In order to understand why we choose this approach, we can imagine an ensemble of three classifiers --- $\{H_1, H_2, H_3\}$ --- and a new text for detection, $x$. If the three classifiers are identical, then when one of them is wrong, they are all going to be wrong together. But if their errors are uncorrelated (i.e., they make different mistakes on different data points), then when $H_1(x)$ is wrong, $H_2(x)$ and $H_3(x)$ may still be correct, and in this case a majority vote will correctly classify $x$.\cite{Agarwal_2023}

A majority vote provides higher confidence than using only one vote, but a unanimous voting strategy, which requires agreement between the three classifiers for a valid prediction, can yield even higher confidence. This might decrease the true-positives rate (TPR), but will ensure the false-positive rate is as low as possible (mistakenly attributing a text to an author).\cite{Agarwal_2023}

The statistical justification is as follows: if the error rates of $L$ hypotheses (classifiers) $H_n$ are all $p$ (where $p < 0.5$), and their errors are at least somewhat uncorrelated, then the probability that the majority vote will be wrong is given by the area under the binomial distribution—that is, the probability that more than $\frac{L}{2}$ classifiers are wrong (for an odd number of classifiers).\cite{dietterich2000ensemble} In unanimous voting, this error rate is even lower, as the ensemble will be wrong only if all classifiers are wrong. Assuming independent errors, the probability for a unanimous error is $p^L$, which represents an exponential decrease in the error probability.\cite{Agarwal_2023}

As discussed earlier, the cost-sensitive nature of this task makes the unanimous voting strategy the preferred method.

\subsection{Model Architecture and Training}
To achieve as much diversity as possible between the classifiers, we choose three classifiers with different architectures that are able to capture different elements in the text.

To be diverse even further, we feed the classifiers with random training data from each of the following LLM-families: OpenAI, Llama, Claude, and Gemini. In order to keep the training set balanced, each classifier receives the same number of training examples from each target family. All data is in the English language.

\subsection{Test Results}
Evaluation of the model is done on a held-out test set that is composed of 50,000 texts from each model family, for a total of 200,000 texts. Since this is a multi-class classification task, and in order to support our cost-sensitive classification task, we use a confusion matrix, precision, recall, and F score for each class with $\beta = 0.5$ (which prioritizes precision over recall), as well as an overall false-positive rate (FPR) and macro-averaged F score with $\beta = 0.5$.

\noindent \textbf{False-Positive Rate (FPR)} in our multiclass setting is defined for each class \(i\) as the ratio of the number of false positives for that class, to the total number of negatives (false positives plus true negatives) for that class:
\[
\text{FPR}_i = \frac{\text{False Positives for class }i}{\text{False Positives for class }i + \text{True Negatives for class }i}.
\]
This quantity represents the proportion of non-\(i\) instances that are incorrectly classified as class \(i\). An overall (macro-averaged) FPR is then computed by averaging the FPRs across all classes:
\[
\text{Overall FPR} = \frac{1}{K} \sum_{i=1}^{K} \text{FPR}_i,
\]
where \(K\) is the number of classes.

\noindent \textbf{Macro-Averaged F Score} is calculated for each class (using $\beta=0.5$ to prioritize precision over recall) and then averaged across all classes.

\subsubsection{Individual Classifiers Metrics}
\begin{table}[ht]
\centering
\caption{Performance of Classifier I}
\label{tab:performance-classifierI}
\begin{tabular}{lcccc|ccc}
\hline
\textbf{True/Predicted} & Claude & Gemini & Llama & OpenAI & Precision & Recall & F$\beta$(0.5) \\ \hline
Claude  & 49083 & 24    & 570   & 323   & 0.9962  & 0.9817 & 0.9933 \\
Gemini  & 36    & 49406 & 224   & 334   & 0.9967  & 0.9881 & 0.9950 \\
Llama   & 22    & 22    & 49687 & 269   & 0.9645  & 0.9937 & 0.9702 \\
OpenAI  & 127   & 116   & 1033  & 48724 & 0.9813  & 0.9745 & 0.9800 \\ \hline
\end{tabular}
\\[1ex]
\small Macro-averaged F$\beta$(0.5) = 0.9846, \quad Macro-averaged FPR = 0.0052
\end{table}

\begin{table}[ht]
\centering
\caption{Performance of Classifier II}
\label{tab:performance-classifierII}
\begin{tabular}{lcccc|ccc}
\hline
\textbf{True/Predicted} & Claude & Gemini & Llama & OpenAI & Precision & Recall & F$\beta$(0.5) \\ \hline
Claude  & 49704 & 10   & 112   & 174   & 0.9981  & 0.9941 & 0.9973 \\
Gemini  & 21    & 49665& 92    & 222   & 0.9983  & 0.9933 & 0.9973 \\
Llama   & 7     & 18   & 49784 & 191   & 0.9895  & 0.9957 & 0.9907 \\
OpenAI  & 67    & 55   & 325   & 49553 & 0.9883  & 0.9911 & 0.9888 \\ \hline
\end{tabular}
\\[1ex]
\small Macro-averaged F$\beta$(0.5) = 0.9935, \quad Macro-averaged FPR = 0.0022
\end{table}

\begin{table}[ht]
\centering
\caption{Performance of Classifier III}
\label{tab:performance-classifierIII}
\begin{tabular}{lcccc|ccc}
\hline
\textbf{True/Predicted} & Claude & Gemini & Llama & OpenAI & Precision & Recall & F$\beta$(0.5) \\ \hline
Claude  & 49326 & 21    & 306   & 347   & 0.9940  & 0.9865 & 0.9925 \\
Gemini  & 71    & 48723 & 550   & 656   & 0.9966  & 0.9745 & 0.9921 \\
Llama   & 52    & 37    & 49450 & 461   & 0.9469  & 0.9890 & 0.9550 \\
OpenAI  & 173   & 108   & 1919  & 47800 & 0.9703  & 0.9560 & 0.9674 \\ \hline
\end{tabular}
\\[1ex]
\small Macro-averaged F$\beta$(0.5) = 0.9768, \quad Macro-averaged FPR = 0.0078
\end{table}

\subsection{Individual Classifiers Results Analysis}
The results demonstrate varying performance between the three classifiers, while achieving overall good performance. 

Classifier I (see table~\ref{tab:performance-classifierI}) performs very well ($F_{\beta} > 0.9$) on texts from the Claude and Gemini datasets, but exhibits slightly lower metrics on Llama and OpenAI data. The confusion matrix shows that it tends to misclassify Llama texts as other categories, mainly as OpenAI ($269$ texts), and that OpenAI texts are misclassified mainly as Llama texts ($1033$ texts). In addition, its false-positive rate is not minor ($0.0052$ or 0.52\%).

Classifier II (see table~\ref{tab:performance-classifierII}) shows much stronger and consistent performance across all four classes, with extremely low FPR of $0.0022$. Its main weaknesses are in classifying OpenAI, mainly as Llama ($325$ texts), and Gemini, mainly as OpenAI ($222$ texts), but this is done in low numbers.

Classifier III (see table~\ref{tab:performance-classifierIII}) shows slightly lower metrics compared to the other two classifiers, mainly on Llama and OpenAI data, as well as with a higher overall FPR of $0.0078$. It struggles mainly with OpenAI (tends to classify texts as Llama with $1919$ errors) and Gemini texts (tends to classify as Llama with $550$ errors and OpenAI with $656$ errors).

\subsection{Ensemble Building and Results}
Given that the error rates of each of these classifiers are far below the 50\% threshold required for effective ensembling, as discussed earlier, and since their errors seem to be at least somehow unrelated, we build two kinds of ensembles with the three classifiers above: majority-vote and unanimous-vote. 

\begin{table}[ht]
\centering
\caption{Performance of the Majority-vote Ensemble}
\label{tab:majorityensemble}
\begin{tabular}{lcccc|ccc}
\hline
\textbf{True/Predicted} & Claude & Gemini & Llama & OpenAI & Precision & Recall & F$\beta$(0.5) \\ \hline
Claude  & 49749 & 8     & 89    & 154   & 0.9983  & 0.9950 & 0.9976 \\
Gemini  & 18    & 49706 & 88    & 188   & 0.9985  & 0.9941 & 0.9976 \\
Llama   & 5     & 16    & 49854 & 125   & 0.9910  & 0.9971 & 0.9922 \\
OpenAI  & 62    & 51    & 276   & 49611 & 0.9907  & 0.9922 & 0.9910 \\ \hline
\end{tabular}
\\[1ex]
\small Macro-averaged F$\beta$(0.5) = 0.9946, \quad Macro-averaged FPR = 0.0018
\end{table}

\begin{table}[ht]
\centering
\caption{Performance of the Unanimous-vote Ensemble.}
\label{tab:unanimousensemble}
\begin{tabular}{lccccc|ccc}
\hline
\textbf{True/Predicted} & Claude & Gemini & Llama & OpenAI & no-agreement & Precision$^*$ & Recall$^*$ & F$\beta$(0.5)$^*$ \\ \hline
Claude  & 49428 & 3    & 26   & 26   & 517   & 0.9991  & 0.9989 & 0.9991 \\
Gemini  & 9     & 49360 & 21   & 29   & 581   & 0.9994  & 0.9988 & 0.9993 \\
Llama   & 19    & 8     & 49583 & 23   & 367   & 0.9984  & 0.9990 & 0.9985 \\
OpenAI  & 17    & 19    & 34   & 49282 & 648   & 0.9984  & 0.9986 & 0.9985 \\ \hline
\end{tabular}
\\[2ex]
\begin{tabular}{lc}
\hline
\textbf{Total no agreement} & \textbf{\% no-agreement} \\ \hline
2113 & 1.06\% \\ \hline
\end{tabular}
\\[1ex]
\small Macro-averaged F$\beta$(0.5)$^*$ = 0.9988, \quad Macro-averaged FPR$^*$ = 0.0004
\\[1ex]
\small $^*$ \textit{Calculated after excluding the “no-agreement” category.}
\end{table}

The majority vote ensemble achieves better performance than any individual classifier alone, with a better balance between precision and recall, as reflected in the F$\beta$ score which is improved by 0.11\%. Its FPR is better than that of classifier II by 18.18\%.

To further improve the FPR, we apply an unanimous ensemble that demonstrates superior precision compared to each individual classifier and the majority vote ensemble. It also exhibits a superior FPR which is 4.5 times lower than the majority-vote ensemble (0.0004). 

The unanimous ensemble is much more conservative in its predictions, leading to higher precision, as expected. However, its main downside is that it introduces a new class - “no-agreement”, which represents cases where the classifiers fail to reach a consensus. This new class accounts for 1.06\% of the examples in the test set (2,113 texts out of 200,000).

As explained earlier, our task is cost-sensitive, because wrongly classifying an LLM writer might have severe implications such as intellectual property, legal, and reputational implications. The unanimous voting ensemble provides better protection against false positives, which are more costly in this case than failing to classify the true writer. This makes the unanimous ensemble a more suitable choice for this task, while acknowledging the inherent trade-off of not obtaining a prediction for approximately 1\% of the texts in the test set.

\subsection{Results Summary}

\begin{table}[ht]
\centering
\caption{Performance Comparison of Ensemble Methods}
\label{tab:results-summary}
\begin{tabular}{lcc}
\hline
\textbf{Method} & \textbf{F\(\beta\)(0.5)} & \textbf{FPR} \\ \hline
Classifier III & 0.9768 & 0.0078 \\
Classifier I     & 0.9846 & 0.0052 \\
Classifier II       & 0.9935 & 0.0022 \\
Majority-vote ensemble       & 0.9946 & 0.0018 \\
Unanimous-vote ensemble      & \textbf{0.9988} & \textbf{0.0004} \\ \hline
\end{tabular}
\end{table}

While we observe a moderate improvement in FPR between the best individual classifier - classifier II - and the majority vote, the improvement from majority to unanimous represents about a 78\% relative decrease. This is significant, even if not an exponential decrease as in the idealized scenario described above, which implies that not all errors are uncorrelated. 

In addition to that, it is interesting to note how strong classifier II is on this task, and a good potential direction for future research might be to understand why it is able to capture more stylistic nuances that the other classifiers miss.

\section{Experiment Application on Unseen LLMs}
\subsection{Rationale and Hypothesis}
The unanimous ensemble has demonstrated high performance on texts from the four LLM families (Claude, Gemini, Llama, and OpenAI). In this section, we assess its generalization capabilities on LLMs that are not included in the training data. 

We hypothesize that when presented with texts from unseen LLMs, the ensemble will exhibit a significantly higher rate of “no-agreement” compared to the 1.06\% observed in the test set. There are three main reasons for this expectation:

(1) The ensemble’s classifiers have learned to identify stylistic fingerprints within a limited feature space that was defined by the four training LLMs. Given that the unseen LLMs have different architectures relative to the four aforementioned LLM families, are built from relatively different training data, and use different generation techniques, we expect that they will have stylistic patterns that fall outside of the learned feature space. 

(2) The effectiveness of the unanimous ensemble is based on the assumption that the errors of the individual classifiers are relatively uncorrelated. Therefore, our expectation is that on texts by unseen models, the ensemble's classifiers will exhibit uncorrelated errors which would reduce the likelihood of unanimous agreement.

(3) The ensemble is designed according to the cost-sensitive approach, with high precision and minimizing the false positives. This conservative design is more likely to abstain from a classification than to wrongly attribute a text to an LLM.

\subsection{Experimental Setup}
We evaluate the ensemble on texts generated by four unseen LLMs:
\begin{itemize}
  \item \textbf{phi-4},\cite{abdin2024phi4technicalreport}
  \item \textbf{Grok-1},\cite{grokwebsite}
  \item \textbf{Mixtral-8x7b-instruct-v0.1},\cite{mistral2023mixtral} and
  \item \textbf{DeepSeek-R1}.\cite{deepseekai2025deepseekr1incentivizingreasoningcapability}
\end{itemize}

For this task, we create a pool of prompts from four HuggingFace repositories:
\begin{itemize}
  \item \texttt{MohamedRashad/ChatGPT-prompts},\footnote{\url{https://huggingface.co/datasets/MohamedRashad/ChatGPT-prompts} (accessed 02.02.2025)}
  \item \texttt{vicgalle/alpaca-gpt4},\footnote{\url{https://huggingface.co/datasets/vicgalle/alpaca-gpt4} (accessed 02.02.2025)}
  \item \texttt{Magpie-Align/Llama-3-Magpie-Pro-1M-v0.1},\footnote{\url{https://huggingface.co/datasets/Magpie-Align/Llama-3-Magpie-Pro-1M-v0.1} (accessed 02.02.2025)} and
  \item \texttt{fka/awesome-chatgpt-prompts}.\footnote{\url{https://huggingface.co/datasets/fka/awesome-chatgpt-prompts} (accessed 02.02.2025)}
\end{itemize}

 These repositories include diverse sets of prompts from multiple domains and styles, ensuring a broad demonstration of the LLMs' capabilities and writing style in the English language. We then shuffle the prompts and randomly choose 13,000 prompts, which are then used to generate texts with the above models. We use the identical set of 13,000 prompts for each of the three LLMs, in order to ensure a fair comparison of their text characteristics. For the Mixtral and DeepSeek models, we use the TogetherAI API for the data generation,\cite{togetherAPI2025} and the phi-4 and Grok-1 models are run locally (as we did not find an inference provider that supports these models).

 Then we input these texts into our LLM-detection ensemble, which produces a “prediction” only when there is unanimous agreement between the three classifiers; otherwise, the text is marked as a “no-agreement”.

 \subsection{Results}
\begin{figure}[ht]
  \centering
  \begin{subfigure}[b]{0.49\linewidth}
    \centering
    \includegraphics[width=\linewidth]{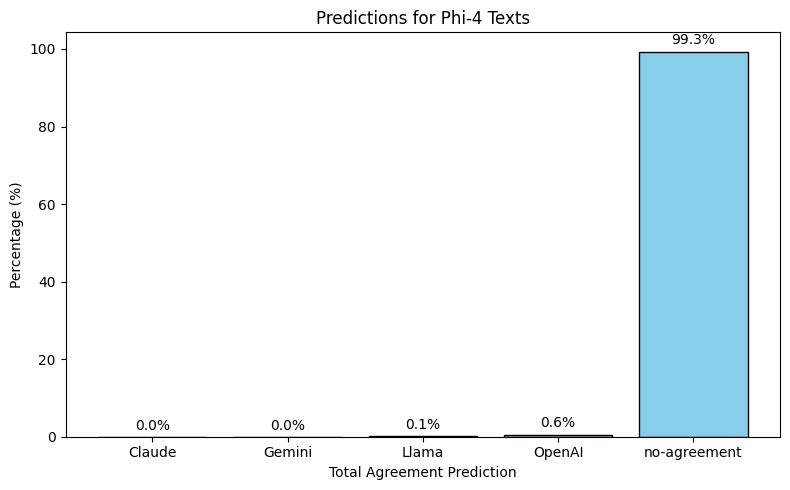}
    \caption{Ensemble results on phi-4 texts.}
    \label{fig:phi4}
  \end{subfigure}
  \hfill
  \begin{subfigure}[b]{0.49\linewidth}
    \centering
    \includegraphics[width=\linewidth]{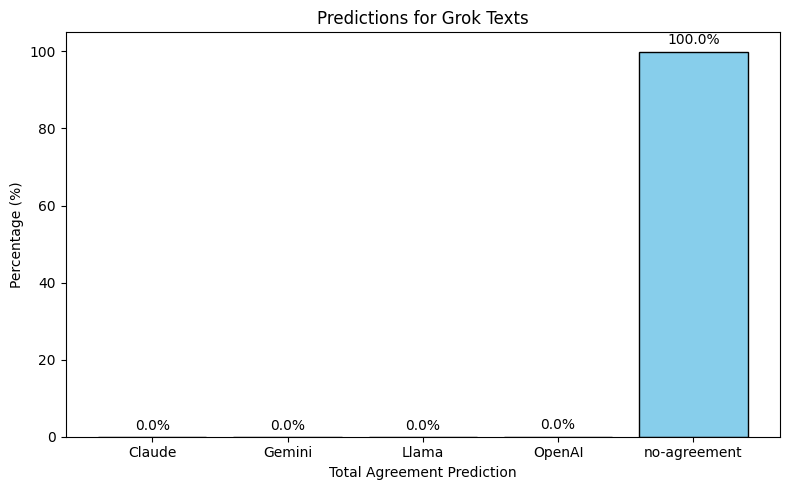}
    \caption{Ensemble results on Grok-1 texts.}
    \label{fig:grok}
  \end{subfigure}
  
  \vspace{1em} 
  
  \begin{subfigure}[b]{0.49\linewidth}
    \centering
    \includegraphics[width=\linewidth]{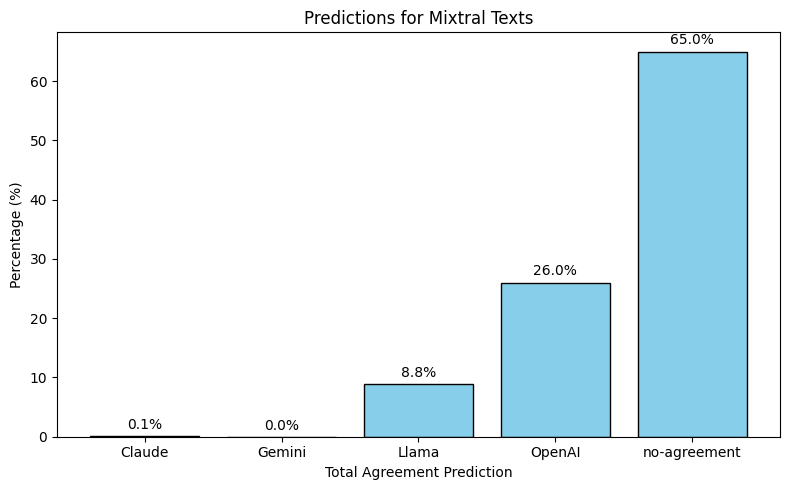}
    \caption{Ensemble results on Mixtral-8x7b-instruct-v0.1 texts.}
    \label{fig:mixtral}
  \end{subfigure}
  \hfill
  \begin{subfigure}[b]{0.49\linewidth}
    \centering
    \includegraphics[width=\linewidth]{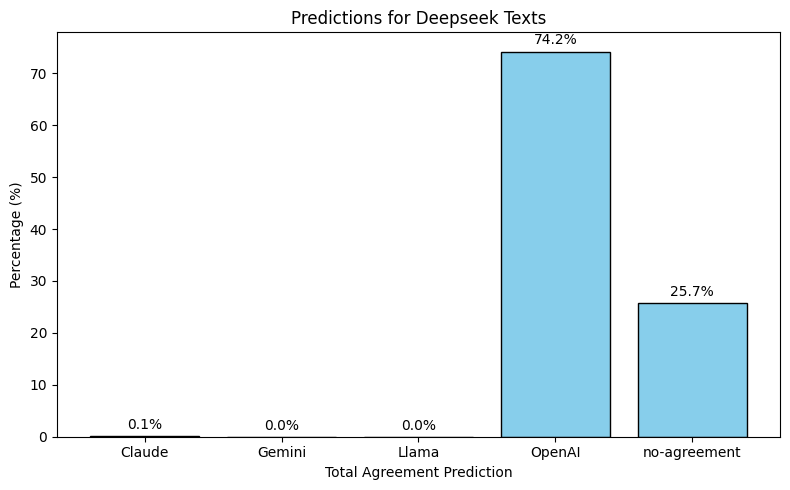}
    \caption{Ensemble results on DeepSeek-R1 texts.}
    \label{fig:deepseek}
  \end{subfigure}
  
  \caption{Comparison of ensemble results from different models.}
  \label{fig:models}
\end{figure}

 For texts by the phi-4 model, as shown in Figure~\ref{fig:phi4}, the ensemble did not agree on 99.3\% of the texts, and for texts by the Grok-1 model, as shown in Figure~\ref{fig:grok}, the ensemble did not agree on 100\% of the texts. This means that the ensemble's classifiers couldn’t agree that the texts by phi-4 or Grok-1 are similar to a particular known LLM writer out of the four LLM-families that this ensemble is trained on.

\begin{figure}[ht]
  \centering
  \includegraphics[width=1\linewidth]{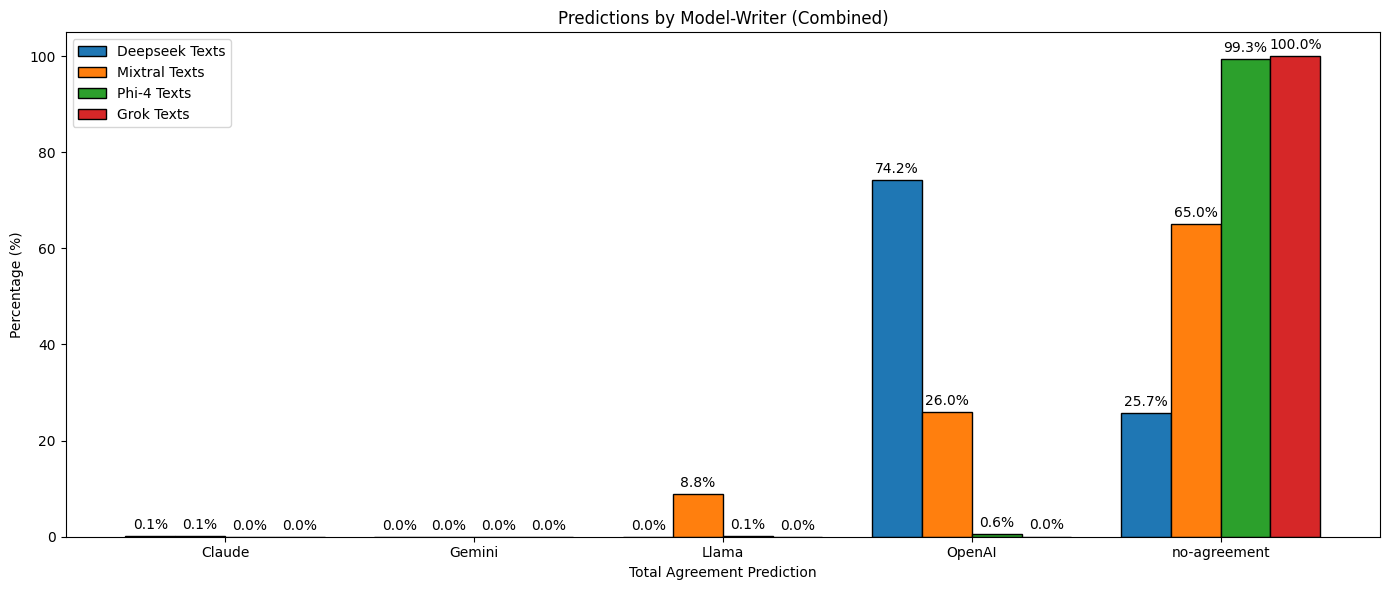}
  \caption{Summary: Ensemble results on four models' texts.}
  \label{fig:summary}
\end{figure}

For texts generated by the Mixtral model, as shown in Figure~\ref{fig:mixtral}, our ensemble did not reach agreement on 65\% of the texts; 26\% were identified as written by OpenAI, and 8.8\% as written by Llama.

For the DeepSeek texts, as shown in Figure~\ref{fig:deepseek}, the ensemble identified 74.2\% of the texts as written by OpenAI, while "no-agreement" was marked for the remaining 25.7\% of the texts. 

The results (see summary in Figure~\ref{fig:summary}) confirm our initial hypothesis that the unanimous ensemble will exhibit a significantly higher rate of “no-agreement” classifications when applied to texts from unseen LLMs, compared to the 1.06\% observed in the test set. This suggests that the ensemble effectively distinguishes between texts generated by LLMs it was trained on, and texts from unseen LLMs.

The extremely high “no-agreement” rate of 99.3\% for the phi-4 model and 100\% for the Grok-1 model, suggests that these models have stylistic fingerprints that are very distinct from the four trained LLMs. These results support the idea that the feature space of our ensemble is defined by the training LLMs, and that LLMs outside this space will be recognized as “no-agreement”.

The 65\% “no-agreement” rate for the Mixtral model suggests that this model possesses mainly distinct stylistic characteristics that are not fully captured by the ensemble. However, the fact that 26\% of the texts were classified as OpenAI and 8.8\% as Llama suggests that there are some stylistic similarities between Mixtral and these LLMs.

The result for the DeepSeek-R1 model is striking. The fact that 74.2\% of the texts were classified as OpenAI suggests strong stylistic similarity between these two LLMs.

\section{Conclusion}
In this work, we introduced a novel and highly cost-sensitive method for classifying AI-generated texts according to their generating LLMs. Our approach involved building a unanimous ensemble of three diverse classifiers trained on texts generated by four LLM families: Claude, Gemini, Llama, and OpenAI. This ensemble achieves extremely high precision (0.9988) coupled with an extremely low false-positive rate (0.0004). Moreover, this ensemble successfully demonstrates its ability to distinguish between texts generated by seen and unseen LLMs.

The application of the ensemble to unseen LLMs exposed interesting insights into the stylistic relationships that exist between models. The high “no-agreement” rates for the phi-4, Grok-1, and Mixtral models, suggest that these models possess unique stylistic features that are not fully captured by the trained ensemble's classifiers. On the other hand, the strong stylistic affiliation between DeepSeek-R1 and OpenAI, as was revealed by the ensemble, raises questions regarding its causes and potential implications.

This work contributes significantly to the area of AI-generated text verification, a contribution which is important to protect intellectual property, to promote transparency and fair usage of AI, to track the evolution of LLMs, and to foster fairness in writing.

Our approach is not without limitations, and these should be addressed in future research. One main aspect omitted from this research is the investigation of the specific linguistic features and fingerprints that contribute to the predictions attributed to the different LLMs. Furthermore, we recommended to include more LLMs for classification, more types of classifiers, more languages, and to explore more kinds of ensemble methods. These can lead to even more explainable and generalizable methods for LLM authorship attribution.

\bibliographystyle{IEEEtran}  
\bibliography{references}

\end{document}